\documentclass[conference]{ieeeconf}  

\IEEEoverridecommandlockouts                              
\usepackage{graphicx} 
\usepackage{subfigure}
\usepackage{float}
\usepackage{caption}
\usepackage{bm}
\usepackage{hyperref}
\usepackage[utf8]{inputenc}
\usepackage[font=small,labelfont=bf]{caption}
\usepackage[
  separate-uncertainty = true,
  multi-part-units = repeat
]{siunitx}
\usepackage[belowskip=-15pt,aboveskip=0pt]{caption}
\title{\LARGE \bf
 Design of Soft, Modular Appendages for a Bio-inspired Multi-Legged Terrestrial Robot
}

\author{Abu Nayem Md. Asraf	Siddiquee, Benjamin Colfer, Yasemin Ozkan-Aydin,\textit{ Member, IEEE}
\thanks{$^{}$This work was not supported by any organization. All the authors are with the Department of Electrical Engineering, University of Notre Dame, Notre Dame, IN 46556 USA
        {\tt\small asiddiq3,bcolfer,yozkanay@nd.edu}}
}

\begin{document}
\maketitle
\thispagestyle{empty}
\pagestyle{empty}
\begin{abstract}
Soft robots have the ability to adapt to their environment, which makes them suitable for use in disaster areas and agricultural fields, where their mobility is constrained by complex terrain. One of the main challenges in developing soft terrestrial robots is that the robot must be soft enough to adapt to its environment, but also rigid enough to exert the required force on the ground to locomote. In this paper, we report a pneumatically driven, soft modular appendage made of silicone for a terrestrial robot capable of generating specific mechanical movement to locomote and transport loads in the desired direction. This two-segmented soft appendage uses actuation in between the joint and the lower segment of the appendage to ensure adequate rigidity to exert the required force to locomote. A prototype of a soft-rigid-bodied tethered physical robot was developed and two sets of experiments were carried out in both air and underwater environments to assess its performance.  The experimental results address the effectiveness of the soft appendage to generate adequate force to navigate through various environments and our design method offers a simple, low-cost, and efficient way to develop terradynamically capable soft appendages that can be used in a variety of locomotion applications. 

\end{abstract}
\section{Introduction}
Developing a terrestrial soft robot that can adapt to harsh unpredictable terrains with high multimodal locomotion efficiency is critical. Extended navigation range, maneuverability, and unprecedented adaptability to unstructured environments make soft terrestrial robots a promising candidate to impart sophisticated functions that are unattainable by traditional rigid-bodied robots \cite{ chen2021legless, lu2018bioinspired}.

The body structure and locomotion mechanics of different biological systems (e. g. cheetah, frog, kangaroo, caterpillar, inch-worm, ant, etc.) \cite{chen2021legless,lu2018bioinspired, huang2018chasing, mao2022ultrafast,wu2019insect,huang2022insect,lin2013soft,shah2021soft,ozkan2021self,waynelovich2016versatile, tang2020leveraging, guo2017design} have been studied and analyzed to develop versatile high-speed soft robots \cite{calisti2017fundamentals} that can achieve multimodal locomotion. 

\begin{figure}[!t]
    \centering
\includegraphics[width=0.5\textwidth,height=0.7\textheight,keepaspectratio]{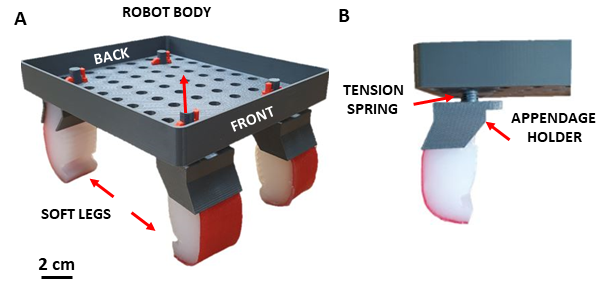}
     \caption{\textbf{Modular, soft-legged quadruped robot.} The robot (mass = 200 gr.) consists of a perforated, rectangular, 3D-printed body part (length = 15.5 cm, width = 13 cm, height = 8.5 cm) and four fabric-reinforced, soft legs that are attached to the perforated body via a spring latch mechanism. }
     \label{fig:robotParts}
\end{figure}
Soft biological systems mostly tend to be small in size as they need to support their own weight without a skeleton, which is considered one of their important limitations. Besides, some soft invertebrates (e.g. giant earthworms, squid, and jellyfish) are found underground or underwater where their body weight is supported by the surrounding solid/liquid mediums \cite{kim2013soft}. To tackle this issue, while developing soft robots, researchers introduced soft-rigid robots \cite{ mao2022ultrafast, shah2021soft, waynelovich2016versatile, tang2020leveraging,guo2017design,xia2021legged,mueller2019voxelated} where the body weight of the soft-rigid robots are supported by rigid parts which allows them to locomote with a higher speed.

Inherent softness and body compliance of soft robots give them the flexibility to achieve infinite degrees of freedom, however, these salient features intensify the level of complexity in the design and control of the soft robot \cite{rus2015design}. Finding the best optimal design, control technique, and material selection are the key factors to develop a soft robot. 

\begin{figure}[!t]
\setlength{\textfloatsep}{0.7\baselineskip plus 0.2\baselineskip minus 0.5\baselineskip}
    \centering
\includegraphics[width=0.5\textwidth,height=0.9\textheight,keepaspectratio]{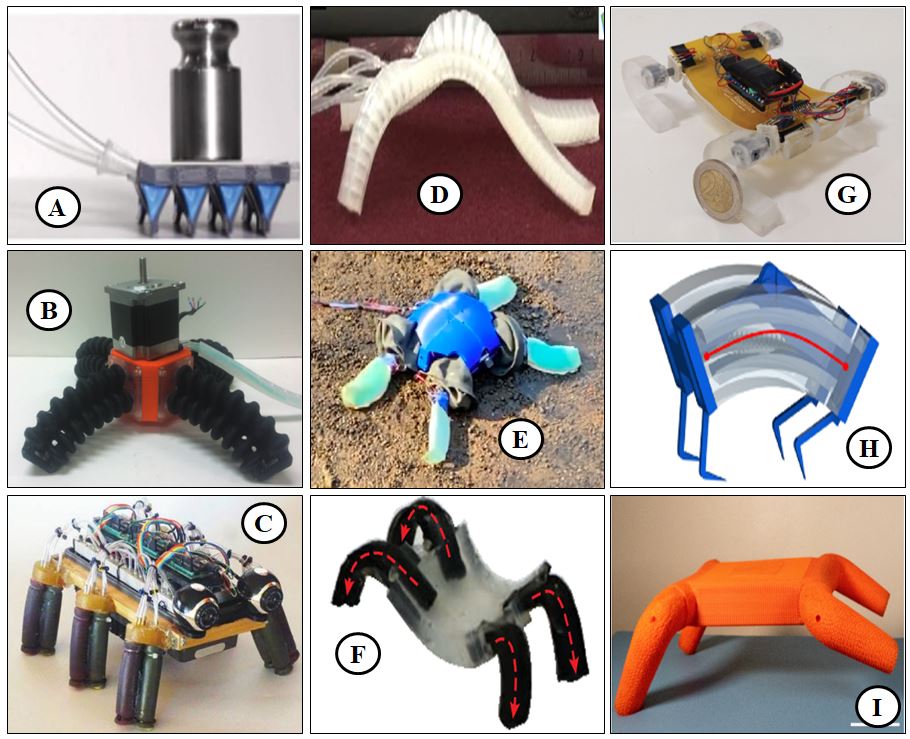}
     \caption{\textbf{Soft terrestrial robots.} Examples of soft, soft-rigid, flexible robots designed for multi-modal locomotion pneumatically driven multi-legged walker (A)\cite{mueller2019voxelated}, (B)\cite{drotman20173d}, (C)\cite{waynelovich2016versatile}, (D)\cite{shepherd2011multigait}, soft-rigid walker and swimmer (E)\cite{baines2022multi}, soft walker (F) \cite{venkiteswaran2019bio}, soft-rigid walker and jumper (G)\cite{kalin2020design},
     (H)\cite{tang2020leveraging},
     (I) \cite{xia2021legged}. } 
     \label{fig:robotexmples}
\end{figure}

An efficient way to develop a complex terrestrial robot is to make it modular which enables them to adapt to the unstructured real environment and simplifies hardware modifications and maintenance  \cite{Yim2007,MOUBARAK20121648,Post2021}. Terrestrial modular legged robots  made mostly by rigid components \cite{drotman20173d,ozkan2021self,OzkanAydin_Centipede} can successfully navigate various challenging terrain.  Besides, there are soft modular robots (Fig.\ref{fig:robotexmples}) which are solely made by soft components, however, their limited control, actuation, and locomotion capabilities have hampered their potential for complex real-world tasks.  Hybrid robots \cite{waynelovich2016versatile,OzkanAydin_Centipede} which have a mix of soft parts as well as rigid parts combine the advantages of both systems and can be more versatile and adaptable.

\subsection{Robot Fabrication}
  \begin{figure*}[!ht]
    \centering
\includegraphics[width=1\textwidth,height=0.7\textheight,keepaspectratio]{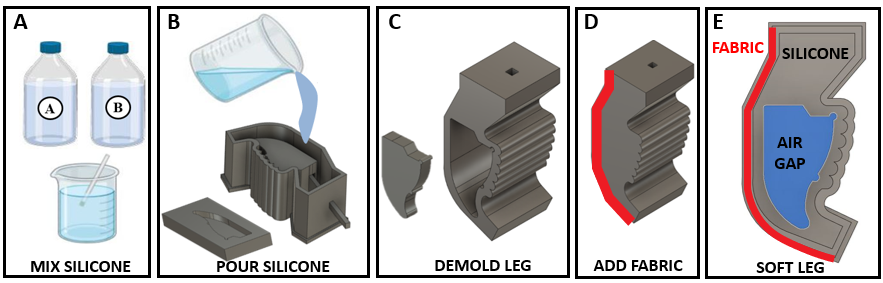}
     \caption{\textbf{Fabrication steps of a Soft Appendage} \textbf{A.} Mix silicone resin (Ecoflex 30,Smooth-On Inc.) 50\% Part A, 50\% Part B, \textbf{B.} Pour silicone resin into the upper and lower parts of the mold, \textbf{C.} Released the cured silicone parts from the mold after 4 hours and join upper and bottom parts, \textbf{D.} Attach fabric layer (red) to the back side of the actuator, \textbf{E.} Fabricated soft leg.}
       \label{fig:fabricationsteps}
\end{figure*}
Although soft body parts have several advantages over their rigid counterparts, the range of exerted force by soft appendages is pre-defined and mostly depends on the design of the leg, applied air pressure, and the property of the material it’s made of. Therefore, increasing the capability of a soft robot to locomote or carry loads is impossible without adjusting its’ structure or material, which is time-consuming and expensive and may not be viable in every circumstance. \\

In this study, we have developed a bio-inspired hybrid modular quadruped robot (Fig. \ref{fig:robotParts}) that utilizes soft appendages to locomote in terrestrial and aquatic environments. We first present the fabrication steps of the robot and soft legs and test its performance in two different environments. We demonstrate that although the robots have the same appendage type and nominal geometry, the orientation of the appendages on the body may lead to discrepancies in the robot's behavior. 

\section{Materials and Methods}

The robot has a 3D-printed, perforated plastic rigid body and four identical soft legs (Fig. \ref{fig:robotParts}). With the four appendages attached the total mass of the robot is approximately 200 gr. and the height, length (body length) and width of the robot are 8.5 cm, 15.5 cm, and 13 cm respectively. The soft legs are connected to the rigid body by appendage holders that are attached to the body via a spring-latch mechanism. This mechanism consists of a tension spring and a connector key that allows the leg to attach to the body tightly. The cylindrical shape of the upper part of the appendage holder allows the appendages to rotate along the z-axis. The perforated structure of the body and the spring-latch connection mechanism allows us to modify the structure of the robot enabling us to easily change the shape of the robot, allowing us to adapt to different environments and tasks.

\subsection{Soft-Leg Fabrication}
We designed and developed a simple version of two segmented (thigh and leg) appendages (height = 6 cm, width = 2.5 cm, mass = 28 gr.) using the epoxy-resin fabrication technique described in \cite{shepherd2011multigait}.  Our aim is to keep the thigh fixed to the ground during locomotion and move the leg by actuating the knee joint.  



Autodesk Fusion-360 (Student version) was used to design the mold of the soft actuator and the robot body. All the molds and the rigid parts of the robot were 3D printed by Stratasys F170 using ABS material. The leg mold has three different parts upper, middle, and lower (Fig. \ref{fig:fabricationsteps}B).  Platinum cure silicon rubber compound, EcoflexTM 00-30, was used to develop the prototype of the soft actuator. EcoflexTM 00-30 comes with two different parts; parts A and B. The same amount of liquid rubber content was taken from parts A and B (volume basis) into a jar and mixed thoroughly to remove extra entrapped air from the mixture (Fig. \ref{fig:fabricationsteps}A). Then the mixture was poured into the upper part of the mold and the middle part of the mold was placed in it and left for 12 hours to be cured in solid (Fig. \ref{fig:fabricationsteps}B). The temperature was maintained at 73°F. After 12 hours of curing time, the top part of the soft actuator was released from the upper part of the mold (Fig. \ref{fig:fabricationsteps}C). Then a uniform mixture of liquid rubber compound of parts A and B is poured into the lower part of the mold. After releasing the lower part it was attached to the upper part of the soft actuator using a silicone mixture.

\begin{figure}[!t]
    \centering
\includegraphics[width=0.25\textwidth,height=0.9\textheight,keepaspectratio]{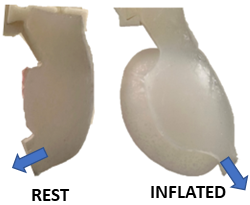}
     \caption{\textbf{Leg actuation states.} Unactuated (left) and actuated (right) states of a leg in the air. The blue arrows show the direction of the toe. }
       \label{fig:legInflate}
\end{figure}

The developed soft actuator needed to follow one more fabrication step. The thickness of the sides of each appendage (5 mm) was chosen in such a way that the leg does not inflate sideways. It was allowed to inflate in its front region. To restrict the actuator's longitudinal elongation, and expansion of the back side, as a strain-limiting layer, a single piece of 100\% pure cotton fabric was wrapped (Fig.\ref{fig:fabricationsteps}D and Fig.\ref{fig:legInflate}) on the back side of the soft actuator to confirm the actuator's unidirectional bending \cite{xavier2021finite, hwang2015pneumatic, konishi2001thin}. 

\begin{figure}[!t]
    \centering
\includegraphics[width=0.5\textwidth,height=0.9\textheight,keepaspectratio]{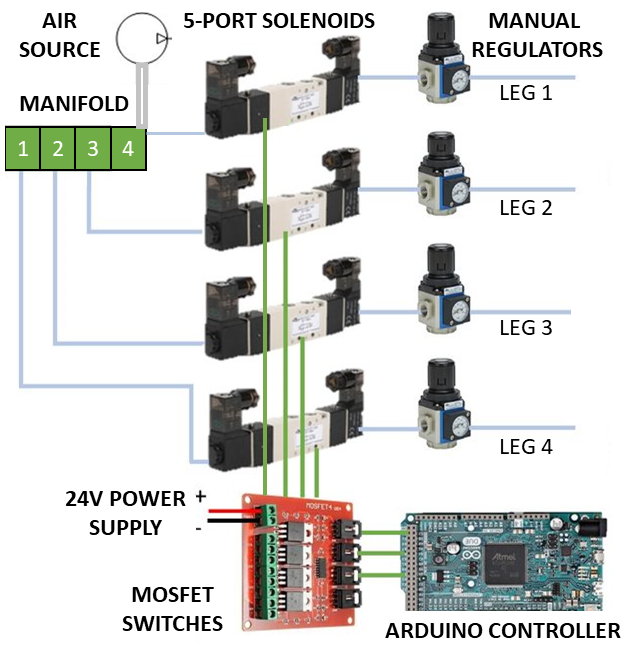}
     \caption{\textbf{Pneumatic Control Board.} The control board includes Arduino Mega controller, four solenoid valves (AVS-513C1-24D, NITRA, 5-port, 4-way, 3-position), four-channel MOSFET switches, four manual pneumatic regulators (AR2-213, NITRA), a manifold (MRA-4CB, NITRA) and an air source}
       \label{fig:pneumaticontrolboard}
\end{figure}
\subsection{Experimental Setup}
We performed two sets of experiments: in the air and underwater. We recorded the experiments using two webcams (Logitech 920) from the side and top view. The videos were synchronized using Logitech Webcam Software. The video frames were extracted at 2 fps and are analyzed using Matlab R2021a. 

In the air, we tested the robot on a flat piece of cardboard. The underwater environment utilized a tank (length = 51 cm, width = 27 cm, height = 30 cm) filled 70\% with water (h = 18 cm). Since the robot body is made of lightweight plastic parts and soft appendages (made of silicone) have hollow actuation chambers of volume 9.3 cm$^3$ in each, the robot was found to float in water during underwater experiments. To prevent the floating of the robot and increase stability (defined as the ability to move on a straight line) we added mass (340 grams) to the robot body. Even though the holes on the top part of the robot were created to make the robot modular so that we can attach/detach appendages according to our requirements, this perforated body helps to reduce the buoyancy exerted on the robot underwater. 

\subsection{Gait Control}
\begin{figure*}[!t]
    \centering
\includegraphics[width=0.8\textwidth,height=0.9\textheight,keepaspectratio]{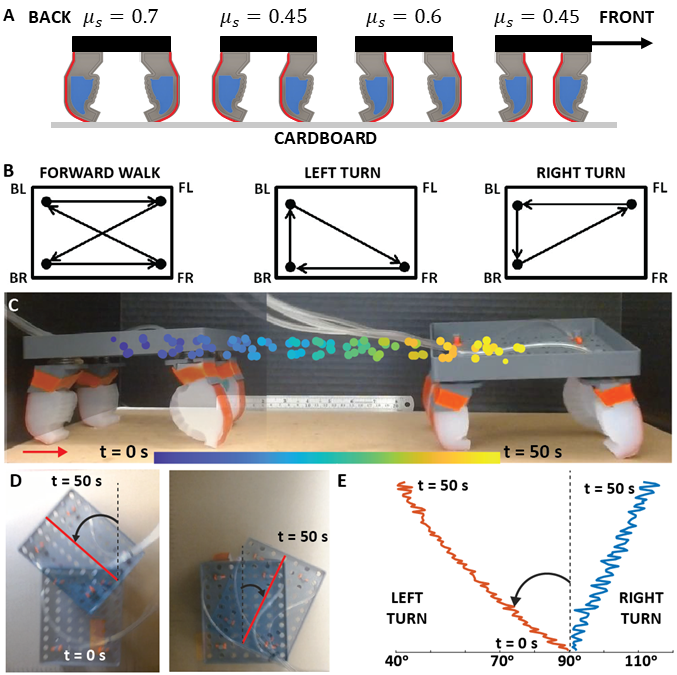}
     \caption{\textbf{Walking and Turning Gaits.  A.} The static friction of the robot on cardboard for different configurations of front and back legs. \textbf{B.} The actuation sequence of legs during forward(left), left turning (middle), and right turning (right). Here FR, FL, BR, and BL indicates the Front Right, Front Left, Back Right, and Back Left appendages, respectively. \textbf{C.} CoM trajectory of the robot in x-z plane when walking on cardboard with a forward gait. Colorbar represents the time (0-50 s.). \textbf{D.} Top view images of the robot when it turns left and right on cardboard. The red line shows the vertical center line of the robot. \textbf{E.} Absolute left and right rotation of the robot after 50 sec for the experiments given in D.}
       \label{fig:wakingturninggait}
\end{figure*}

An Arduino-controlled pneumatic control board (Fig. \ref{fig:pneumaticontrolboard}) was used to control the gaits of the robot. According to the sequence of the actuation of appendages given in Fig.\ref{fig:underwaterwalking}B, three algorithms were developed using Arduino IDE. Each appendage was connected to an Arduino Mega-controlled pneumatic pressure control board that includes four solenoid valves (AVS-513C1-24D, NITRA, 5-port, 4-way, 3-position) controlled via four-channel MOSFET switches, four manual pneumatic regulators (AR2-213, NITRA), a manifold (MRA-4CB, NITRA), and a 24V power supply. The wall air supply source is connected to the control board and before the actuation of the legs the pressures of each actuator were manually set to a max pressure that the legs can tolerate without blowing via manual regulators.  Activating the solenoids with sequences given in Fig.\ref{fig:wakingturninggait}B, we obtained forward walking and turning (left or right) gaits. 
\section{Experimental Results and Discussion}

Our experiments aim to assess the performance of the robot in terrestrial and aquatic environments and measure the effect of different leg configurations on the robot’s mobility and maneuverability. We performed three experiments per each gait type starting from the same initial conditions and calculated the mean and standard deviation of the measurements (displacement and rotation of the body) using Matlab R2021a.

Figure \ref{fig:wakingturninggait}-A gives the static friction measurements on cardboard due to different orientations of soft appendages. To measure the static friction, we put the robot onto the cardboard and pulled it horizontally with a non-extendible wire using Nextech DFS100 force gauge for four different leg configurations. The appendages need to overcome maximum friction force when the back and front appendage pairs are facing each other. In our case the robot needs more friction from the ground, otherwise, the bottom part of the appendage will not be able to exert adequate force on the ground when actuated rather it will slide due to low friction (supplementary movie). To generate a higher positive resultant force after striking the surface we attached the pair of back and front appendages facing each other, which has a maximum static friction constant ($\mu_s=0.7$). 

Figure \ref{fig:wakingturninggait}-B illustrates the sequences of appendage actuation for forward movement (left), left (middle), and right turning (right). Here FR, FL, BR, BL indicate the \textbf{F}ront \textbf{R}ight, \textbf{F}ront \textbf{L}eft, \textbf{B}ack \textbf{R}ight and \textbf{B}ack \textbf{L}eft appendages, respectively. 

Figure \ref{fig:wakingturninggait}-C shows the robots’ trajectory during forward walking on cardboard (X-Z plane). Here, the color bar indicates the time, and colored markers indicate the center of mass (CoM) position of the robot as a function of time. The robot completed each gait cycle in $\sim$4 sec and successfully traveled the 35 cm in 50 sec.  Figure \ref{fig:wakingturninggait}-D represents the robots’ initial and final positions (at 50 seconds) on the x-y plane and (E) depicts the robots’ motion paths during left and right turns(x-y plane). 

 For forward movement, we actuated the legs in a lateral sequence FR-BL-FL-BR \cite{Hildebrand}. Air was injected in each appendage for 0.22 seconds to inflate followed by a delay of 0.75 seconds to deflate. The duration of a complete cycle was 3.88 seconds; therefore, each appendage was under actuation for 5.67\% time of a complete cycle. During the forward movement, the speed of the robot was $0.21 \pm 0.2$ Body length per cycle (BL/cycle) and the robot was found to oscillate approximately 1.1 cm in the vertical plane (Figure \ref{fig:wakingturninggait}-C).
 
\begin{figure*}[!t]
    \centering
\includegraphics[width=1\textwidth,height=0.9\textheight,keepaspectratio]{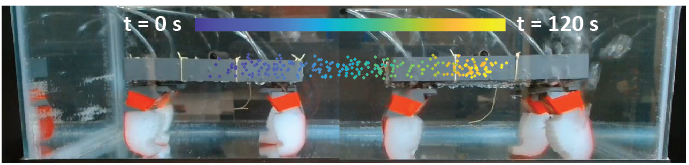}
     \caption{\textbf{Underwater walking.}CoM trajectory of the robot in x-z plane when walking underwater with a forward gait. Colorbar represents the time (0-120 s.). Air bubbles come out of the actuators because of manufacturing errors.}
     \label{fig:underwaterwalking}
\end{figure*}

The actuation sequence of the appendages for turning left was FR-BR-BL; where the duration of a complete cycle was 1.62 seconds. To make the robot turn left, the appendages were inflated for 0.09 seconds, which is the 5.55\% of the time of a complete cycle and 27.77\% time of a complete cycle or 0.45 seconds to deflate the appendages. The average rotational speed of the robot was $1.62^{\circ}\pm 0.05^{\circ}$/cycle while turning left (Fig. \ref{fig:wakingturninggait}-D ). For turning right, the appendages were actuated for 0.15 seconds (6.41\% time of a complete cycle) followed by 0.63 seconds (26.92\% time of a complete cycle) delay in the sequence of FL-BL-BR. Here, the duration of a complete cycle to turn right was 2.34 seconds.  During turning right the robot’s measured rotational speed was $1.4^{\circ}\pm 0.04^{\circ}$/cycle see Figure \ref{fig:wakingturninggait} (D). Although the sequences are reversed when turning left and right, the actuation time of the legs needs to be slightly different to overcome manufacturing errors.

Figure \ref{fig:underwaterwalking} shows the robots’ forward motion path in the aquatic environment (X-Z plane). The robot was found to oscillate between 0 to 3 cm while moving forward underwater and the measured speed of the robot was $0.13 \pm 0.03$ (BL/cycle).  

The robot uses three out of four appendages to take left or right turns. During left and right turns, the FL and FR appendages remain unactuated throughout the cycle; but these appendages help the robot maintain stability. From left turn to right turn, the actuation sequence of the appendages was flipped. Therefore, in both the cases for turning left and right, the ratios of actuation times and deflation times were supposed to be the same, but, there is a deviation of 3.8\% in their values. 

The robot was found to oscillate in the Z direction underwater more than that walking on the ground. The perforated body of the robot helps it to attach appendages at any position and at the same time facilitates it to sink underwater. To achieve stability underwater we added extra weight on top of the robot and in the water, the robot had to withstand buoyancy force, lift and drag forces (as the robot was walking), therefore the linear speed of the robot underwater was less than the linear speed of the robot on the ground. 

\section{Conclusions}
In this work, we presented a new soft appendage and developed a low-cost, light-weight, soft, modular multi-legged quadruped robot that is capable of navigating terrestrial and aquatic mediums without changing its structure. We tested the performance of the robot by conducting experiments and showed that the robot successfully locomote in a straight path and  turn left/right. The robot was capable of moving in the forward direction in an aquatic (water) medium using the same gait it used in terrestrial environments, which expands the range of its application area.  

However, there are some limitations that needed to be eliminated to improve the stability and locomotion performance of the robot. The manufacturing errors associated with each of the soft appendages cause them to exert individual forces which are not identical in magnitudes despite actuating them with the same amount of pressure. Due to the softness of the appendage and the air injection tubing, all the appendages were not able to hold the same pressure (air leakage) while inflating and at the same time the appendages were unable to deflate within the same amount of time when the air pressure was released. To facilitate deflation we used needles; and the robot was tethered, which disturbed the robot's stability while moving and impacted on the experimental results.

 Future work could focus on making this robot untethered and autonomous. An effort could be made to optimize the designs of the soft appendage and the robot to improve the robot's performance. Finite Element Method (FEM) simulations can predict the behavior of the soft actuators which can help us to design the appendage according to the requirement. Furthermore, in this work, we  used Ecoflex-30A silicone for developing the soft appendages. There are different types of silicone with different tensile strengths and other material properties. We can explore different materials and find the best candidate for the fabrication of the soft appendage.

\section{Acknowledgments}
We would like to acknowledge Sean Even for building the pneumatic control board and for technical support, and Enes Aydın for his assistance in the experiments.

\bibliographystyle{ieeetr}

\end{document}